\documentclass[letterpaper,journal,10pt]{IEEEtran}
\usepackage{amsmath,amsfonts}
\usepackage{algorithmic}
\usepackage{algorithm}
\usepackage{array}
\usepackage[caption=false,font=normalsize,labelfont=sf,textfont=sf]{subfig}
\usepackage{textcomp}
\usepackage{stfloats}
\usepackage{url}
\usepackage{verbatim}
\usepackage{graphicx}
\usepackage{cite}
\begin{document}

\title{Real-Time Animatable 2DGS-Avatars with Detail Enhancement from Monocular Videos}

\author{Xia Yuan, Hai Yuan, Wenyi Ge, Ying Fu, Xi Wu, Guanyu Xing$^*$

\thanks{
\textit{Corresponding author:Guanyu Xing.}

This work has been submitted to the IEEE for possible publication. Copyright may be transferred without notice, after which this version may no longer be accessible.}
}


\maketitle

\begin{abstract}
High-quality, animatable 3D human avatar reconstruction from monocular videos offers significant potential for reducing reliance on complex hardware, making it highly practical for applications in game development, augmented reality, and social media. However, existing methods still face substantial challenges in capturing fine geometric details and maintaining animation stability, particularly under dynamic or complex poses. To address these issues, we propose a novel real-time framework for animatable human avatar reconstruction based on 2D Gaussian Splatting (2DGS). By leveraging 2DGS and global SMPL pose parameters, our framework not only aligns positional and rotational discrepancies but also enables robust and natural pose-driven animation of the reconstructed avatars. Furthermore, we introduce a Rotation Compensation Network (RCN) that learns rotation residuals by integrating local geometric features with global pose parameters. This network significantly improves the handling of non-rigid deformations and ensures smooth, artifact-free pose transitions during animation. Experimental results demonstrate that our method successfully reconstructs realistic and highly animatable human avatars from monocular videos, effectively preserving fine-grained details while ensuring stable and natural pose variation. Our approach surpasses current state-of-the-art methods in both reconstruction quality and animation robustness on public benchmarks.
\end{abstract}

\begin{IEEEkeywords}
Human Avatar, Animatable, 2D Gaussian Splatting, Detail Enhancement.
\end{IEEEkeywords}

\section{Introduction}
\IEEEPARstart{H}{uman} avatar modeling from monocular video enables the creation of a high-fidelity, animatable human avatar using single-view RGB sequences. This technology significantly reduces reliance on complex multi-view systems, thereby improving both the practicality and accessibility of human avatar modeling. It also offers broad application potential across domains, including video games, augmented reality, digital humans, and social media. However, reconstructing a human avatar from a single monocular video presents several challenges, including limited viewpoint reconstruction, extreme pose animation, and complex surface deformation modeling.

Most existing methods leverage human parameter models, such as SMPL~\cite{loper2015smpl}, as priors to overcome the constraints imposed by the limited perspective. Among them, some works utilize a human mesh provided by the SMPL model, aligning the mesh vertices with image pixels to achieve the reconstruction of human models~\cite{alldieck2018video,alldieck2018detailed}. However, these methods are fundamentally limited in reconstruction accuracy and computational efficiency. Additionally, the fixed topology of parametric models restricts the ability to capture intricate geometric details and realistic dynamic deformations of human avatars. To address topological rigidity, implicit representations have been explored for animatable human avatar reconstruction, such as Signed Distance Function (SDF)-based methods~\cite{huang2020arch,jiang2022selfrecon} and Neural Radiance Fields (NeRF)-based methods~\cite{zheng2022structured,liu2021neural,peng2021animatable,zhang2023explicifying}. Although these approaches are capable of reconstructing high-precision human models, The differentiable deformation and volume rendering of implicit fields, along with global optimization computations, are highly time-consuming, significantly reducing the practicality of such methods. Moreover, the implicit representation itself is not conducive to editing human movements, limiting their applicability for real-time human animation. Fortunately, 3D Gaussian Splatting (3DGS) offers a novel explicit representation of 3D scenes and enables real-time rendering, opening up new possibilities for constructing real-time animated human avatars. Several methods integrate SMPL mesh vertices with learnable 3D Gaussian kernels and utilize forward linear blending skinning (LBS) for animation, significantly improving both the rendering quality and efficiency of human avatars~\cite{shao2024splattingavatar, qian20243dgs, kocabas2024hugs}. However, the anisotropic Gaussian kernels used in 3DGS are ill-suited for capturing the thin, high-curvature surfaces of the human body, making high-precision human geometry reconstruction particularly challenging~\cite{huang20242d}. 

The recently proposed 2D Gaussian Splatting (2DGS) technique~\cite{huang20242d} significantly enhances surface geometry fidelity by replacing anisotropic 3D Gaussian kernels with view-dependent 2D Gaussian primitives. Reconstructing high-quality human avatars with 2D Gaussian primitives requires accurately fitting the dynamic human body surface. In practice, most Gaussian Splatting-based human avatar reconstruction approaches rely solely on local triangle embeddings anchored to SMPL-based priors, which typically model human motion as rigid transformations of mesh triangles~\cite{shao2024splattingavatar}. However, the shape of clothing is influenced by body movements, and significant deformations occur in the joint areas during motion~\cite{kocabas2024hugs}. Obviously, using only rigid transformations like rotation and translation is insufficient to accurately capture the geometric details of the body caused by deformation. Furthermore, LBS-driven SMPL models fail to capture subtle, nonlinear variations in human surface geometry due to coarse joint rotation interpolation, leading to discrepancies with the real human body. Consequently, the 2D Gaussian primitives embedded in SMPL do not accurately adhere to the actual human surface. These issues lead to reconstruction artifacts such as misalignment, floating, and blurred contours, posing a significant challenge in creating high-quality and animatable human avatars. Unfortunately, there is limited research on methods for animated human avatar reconstruction based on 2DGS. Although Zhao et al. propose a 2DGS-based method for human reconstruction~\cite{zhao2024surfel}, their approach primarily focuses on re-lighting the human body, with limited exploration of human modeling and animation.

This paper proposes a real-time, animatable 2DGS-avatar reconstruction framework based on monocular video. The framework dynamically learns the local transformation residuals of 2D Gaussian primitives induced by non-rigid deformations during motion, leveraging both local body features and global SMPL pose parameters, thereby enabling the reconstruction of high-quality human avatars with rich geometric details. To achieve this, we first embed 2D Gaussian primitives into the SMPL mesh and introduce offset vectors to model surface geometric positional residuals. These offsets compensate for displacement discrepancies between the SMPL model and the actual human body, thereby improving the overall reconstruction quality. To better model the complex deformations in joint regions, a greater number of 2D Gaussian primitives are inserted into these areas. Moreover, to address the bias caused by applying LBS to model non-rigid deformations within the 2DGS rotation parameter space, we propose a Rotation Compensation Network~(RCN). RCN learns local rotation residuals by integrating global pose parameters with local attributes, such as triangle index embeddings, the positions of 2DGS primitives, as well as distance and rotation offsets. Finally, we design constraints targeting both the reconstructed human appearance and the geometric attributes of the 2D Gaussian primitives to achieve more accurate human avatar reconstruction and ensure smoother animation transitions.

Experimental results show that our method reconstructs high-quality, animatable human avatars with fine details from monocular video, while ensuring smooth and consistent pose transitions by effectively handling non-rigid deformations and joint artifacts. It outperforms state-of-the-art methods on public benchmarks. Our main contributions are as follows:

\begin{itemize}
    \item We propose a real-time framework for animatable human avatar reconstruction from monocular video. By introducing 2DGS, the framework leverages local body features and global SMPL pose parameters to correct positional and rotational discrepancies between 2DGS primitives and the actual human body surface, which enables the reconstruction of high-quality avatars with rich geometric details.
    \item We design a Rotation Compensation Network (RCN) that learns a rotation residual function by leveraging local geometric features together with global SMPL pose parameters, effectively overcoming the limitations of rigid SMPL embeddings and coarse LBS, and enabling the capture of high-frequency rotational variations while ensuring smooth and precise pose transitions.
    \item Experimental results on public datasets demonstrate that our method outperforms existing state-of-the-art human avatar reconstruction approaches.
\end{itemize}

\section{Related Work}
\label{sec:related}
\subsection{Explicit Representation-Based Human Avatars}
Reconstructing detailed human avatars from a single video remains challenging due to pose variation, complex clothing topology, and monocular depth ambiguity. SCALE~\cite{ma2021scale} uses point clouds to capture overall shape but cannot produce continuous surface detail. By contrast, Mesh representations use vertex and triangle topology to form continuous surfaces, and with UV coordinates for texture mapping, they offer clear advantages in capturing fine details such as clothing wrinkles and appearance. Parametric models like SMPL~\cite{loper2015smpl} provide a powerful, efficient human prior~\cite{yang2024innovative}.
Some methods~\cite{alldieck2018video,alldieck2018detailed,lazova2019360} reconstruct human bodies by offsetting SMPL vertices.Alldieck et al.\cite{alldieck2018video} transform each frame’s silhouette cone into a canonical pose and back-project multi-view colors onto SMPL vertices to build complete texture maps. Lazova et al.\cite{lazova2019360} infer partial UV textures from input views and fuse them via segmentation layouts to generate full-body UV maps for clothed human modeling. While these methods offer flexibility in personalized modeling, the fixed topology of the SMPL model hinders the modeling of more complex and dynamic geometric details of human avatars.

\subsection{Implicit Representation-Based Human Avatars}
Fixed mesh topologies limit complex human avatar modeling. To address this, recent works~\cite{jiang2022selfrecon,huang2020arch,liao2023high,zuo2020sparsefusion} use continuous implicit representations, whose flexibility adapts naturally to varying body and clothing topologies.
Jiang et al. \cite{jiang2022selfrecon} fuse per-frame explicit meshes from a monocular rotating video with a canonical implicit SDF, then self-supervise its geometric optimization via non-rigid ray casting and differentiable rendering. Liao et al. \cite{liao2023high} learn a canonical body shape in an implicit framework and refine fine surface details by estimating non-rigid deformations.
Since Neural Radiance Fields (NeRF) \cite{mildenhall2021nerf} emerged, 3D rendering quality has improved dramatically. By leveraging neural fields' superior function fitting capabilities, complex human geometries can be reconstructed with smooth surfaces. several works~\cite{weng2022humannerf, peng2021neural,peng2024animatable,kwon2025text2avatar,wang2022nerfcap} integrate the SMPL model with neural implicit fields to animate humans. Peng et al. \cite{peng2021neural} anchor latent codes at SMPL vertices and diffuse them via sparse convolutional MLPs to regress density and color for any 3D point. However, deformation fields may overfit the training data due to depth ambiguity and limited pose observations inherent in monocular videos, leading to floating artifacts in novel views. Weng et al. \cite{weng2022humannerf} address this by optimizing canonical pose volumetric and motion fields, splitting motion into skeletal rigid and nonrigid components for high‐fidelity rendering. Despite these advances, neural methods remain computationally heavy, limiting rendering speed. 

\subsection{Gaussian-Based Human Avatars}
Compared to implicit neural representations, 3D Gaussian Splatting (3DGS)\cite{kerbl20233d} represents radiance fields using anisotropic Gaussians with color attributes, enabling real-time, photorealistic rendering via differentiable rasterization. While highly effective for static scenes\cite{wang2024geometry, liu2025citygaussian}, recent extensions to dynamic human avatars~\cite{li2024animatable, jiang2024gs} often rely on costly multi-view capture systems, limiting practical deployment. To mitigate this, monocular methods~\cite{kocabas2024hugs, lei2024gart, moreau2024human, shao2024splattingavatar, hu2024gaussianavatar} utilize SMPL priors and linear blend skinning (LBS) to animate 3D Gaussians, achieving faster novel-view rendering. Shao \emph{et al.}\cite{shao2024splattingavatar} improve spatial accuracy by embedding Gaussians on mesh triangles using barycentric coordinates and displacement offsets, while Hu \emph{et al.}\cite{hu2024gaussianavatar} enhance appearance fidelity through a dynamic network that predicts motion-dependent attributes.
Nevertheless, 3DGS exhibits multi-view inconsistencies, hindering fine detail reconstruction. To address this, 2D Gaussian Splatting (2DGS)\cite{huang20242d} introduces oriented disks in the 3D tangent space along with normal and depth distortion losses, significantly improving geometric precision. Zhao \emph{et al.}\cite{zhao2024surfel} further integrate physically based rendering (PBR) parameters for realistic avatar rendering under varying illumination, though their focus remains on appearance editing. Drawing from these insights, we incorporate 2DGS~\cite{huang20242d} to achieve efficient and detailed modeling of animatable human avatars.

\begin{figure*}[!t]
  \centering
  \includegraphics[width=1\linewidth]{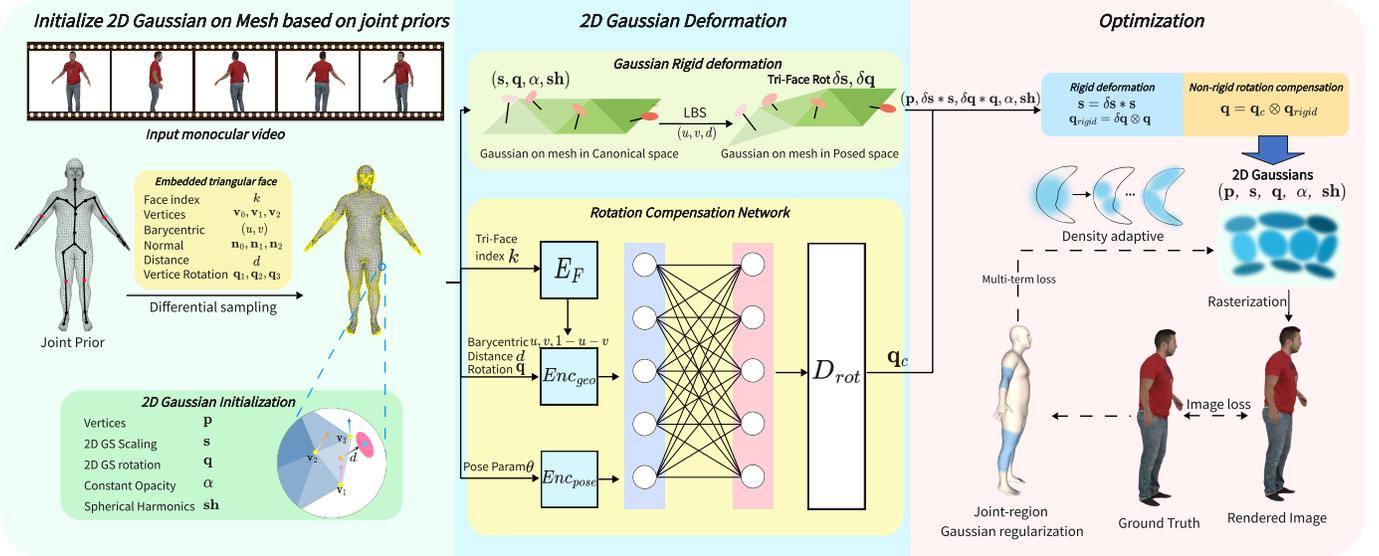}
  \caption{We perform SMPL joint-prior guided non-uniform sampling across the mesh surface. Each 2D Gaussian primitive is embedded into a sampled triangle using barycentric coordinates $(u,v)$ and an offset distance $d$ measured along the triangle’s normal $\mathbf{n}$. We then deform the mesh via Linear Blend Skinning to transfer the 2D Gaussians primitives from the canonical space to the posed space and apply our proposed RCN to refine their orientations. Finally, we jointly optimize the Gaussian embedding parameters and the RCN weights through differentiable rasterization.}
  \label{fig:pipeline}
\end{figure*}

\section{Method}
\label{sec:method}
We propose a novel real-time framework for animatable human avatar reconstruction from monocular video, using 2D Gaussian Splatting (2DGS). The overall structure of the framework is illustrated in Fig.~\ref{fig:pipeline}. The input consists of video frames captured by a monocular camera, showing a human subject performing an in-place rotation while maintaining an initial A-pose. In the initialization stage, we introduce a method to embed 2D Gaussian primitives into the SMPL mesh of the target human subject defined in canonical space~(i.e., the A-pose). Guided by joint priors, we adopt a dense sampling strategy to increase the number of primitives near joint regions, which improves the accuracy of representation in areas that experience large deformations~(Sec.~\ref{sec:embedded}). In the subsequent deformation stage, the embedded primitives are deformed from canonical space to posed space using Linear Blend Skinning (LBS)(Sec.~\ref{sec:Transformations}). To better capture fine non-rigid deformations that cannot be handled by rigid transformations alone, we add a Rotation Compensation Network (RCN), which predicts local rotation offsets using geometric features and global pose parameters~(Sec.~\ref{sec:rotation}). In the optimization stage, we reconstruct the human avatar by jointly adjusting all parameters of the 2D Gaussian primitives. This is done by minimizing the difference between the rendered images and the input frames, while also applying shape constraints to maintain geometric consistency and rendering quality~(Sec.~\ref{sec:loss}). At inference time, new poses are generated by first applying LBS-based coarse deformation with the given pose parameters, followed by refinement using RCN. The following subsections describe the technical implementation of each part of the proposed framework.

\subsection{Preliminary}
\label{sec:Preliminary}
\noindent\textbf{2D Gaussian Splatting~\cite{huang20242d}} reduces one scale dimension of the 3D Gaussians Splatting~\cite{kerbl20233d} to zero, forming surface-aligned Gaussian primitives. This design enhances view consistency. 
2D Gaussian primitives are defined on the local tangent plane of the scene surface. Each primitive is parameterized by a center point $\mathbf{p}_i$, two orthogonal tangential vectors $\mathbf{t}_u$ and $\mathbf{t}_v$ defining the tangent plane orientation, and scaling vectors $(\mathbf{s}_u,\mathbf{s}_v)$ controlling anisotropic scaling along these directions. The normal vector $\mathbf{t}_w$ is derived via cross product $\mathbf{t}_w = \mathbf{t}_u \times \mathbf{t}_v$ and its orientation can be arranged into a $3\times3$ rotation matrix $\mathrm{R} = [\mathbf{t}_u,\, \mathbf{t}_v,\, \mathbf{t}_w]$. The scaling factors can be further organized into a $3 \times 3$ diagonal matrix $\mathrm{S} = \mathrm{diag}(\mathbf{s}_u,\, \mathbf{s}_v,\, 0)$. Each 2D Gaussian primitive is thus defined on a local tangent plane in world space and represented as follows:

\begin{equation}
    P(u,v)=\mathbf{p}_i+\mathbf{s}_u\mathbf{t}_uu+\mathbf{s}_v\mathbf{t}_vv=\mathrm{H}(u,v,1,1)
\end{equation}
\begin{equation}
\mathrm{H} = \begin{bmatrix}
\mathbf{s}_u \mathbf{t}_u & \mathbf{s}_v \mathbf{t}_v & 0 & \mathbf{p}_i \\
0 & 0 & 0 & 1
\end{bmatrix} = \begin{bmatrix}
\mathrm{RS} & \mathbf{p}_i \\
0 & 1
\end{bmatrix}
\end{equation}

where, $\mathrm{H} \in \mathbb{R}^{4 \times 4}$ is a homogeneous transformation matrix representing the 2D Gaussian geometry. For a point $\mathbf{u} = (u, v)$ defined in the local space of a 2D Gaussian primitive, its world-space position can be computed as:
\begin{equation}
    \mathcal{G}(\mathbf{u})=\exp\left(-\frac{u^2+v^2}{2}\right)
\end{equation}

The 2D Gaussian primitive is represented as $G =\{\mathbf{p}_i,\, \mathbf{s},\, \mathbf{q},\, \alpha,\, \mathbf{sh}\}$, where $\mathbf{p}_i$ denotes the center, $\mathbf{q}$ is the rotation, $\alpha$ is the opacity, and $\mathbf{sh}$ represents the view-dependent appearance encoded using spherical harmonics.

During rendering, 2D Gaussian primitives are sorted based on their center depth and organized into tiles. To render a pixel $\mathbf{x} = (x, y)$, a ray is cast into the scene, and the intersection $\mathbf{I(x)}$ with 2D Gaussian primitive is computed. Subsequently, the texture color is accumulated using volumetric alpha blending:
\begin{equation}
    c(\mathbf{x})=\sum_{i=1}c_i\alpha_i\mathcal{G}_i\left(\mathbf{I(x)}\right)\prod_{j=1}^{i-1}\left(1-\alpha_j\mathcal{G}_j\left(\mathbf{I(x)}\right)\right)
\end{equation}

\noindent\textbf{SMPL~\cite{loper2015smpl}} is a widely adopted parametric human mesh model comprising $6890$ vertices, $13776$ triangular faces, and $24$ joints. Each vertex $\mathbf{v}_i$ is associated with a 3D coordinate and a weight vector $\mathbf{w}_i$, where $\mathbf{w}_i$ quantifies the influence of the $24$ joints on that vertex. SMPL deforms the mesh via Linear Blend Skinning (LBS), where pose parameters $\boldsymbol{\theta}$ (joint rotations) and shape parameters $\boldsymbol{\beta}$ (body shape) drive the deformation.
In this work, We assume that the pose and shape parameters for each input video frame are known.

\subsection{Embedding 2D Gaussian Primitives into the SMPL Mesh}
\label{sec:embedded}
To embed 2D Gaussian primitives in canonical space, each primitive is assigned to a specific triangle on the SMPL mesh, defined by its three vertices ${\mathbf{v}_0, \mathbf{v}_1, \mathbf{v}_2}$ and their corresponding normals ${\mathbf{n}_0, \mathbf{n}_1, \mathbf{n}_2}$. For any point $\mathbf{p}_x$ on the triangle, its position and normal $\mathbf{n}_x$ can be interpolated using barycentric coordinates $(u,v)$ as follows:
\begin{equation}
   \mathbf{p}_x=u\mathbf{v}_0+v\mathbf{v}_1+(1-u-v)\mathbf{v}_2
      \label{equ:eq1}
\end{equation}
\begin{equation}
    \mathbf{n}_x=u\mathbf{n}_0+v\mathbf{n}_1+(1-u-v)\mathbf{n}_2
          \label{equ:eq2}
\end{equation}

We compute the center position $\mathbf{p}_i$ of the embedded 2D Gaussian primitive using the point \(\mathbf{p}_x\), its normal \(\mathbf{n}_x\), and the offset distance \(d\), as follows:
\begin{equation}
    \mathbf{p}_i=\mathbf{p}_x+\mathbf{n}_xd
          \label{equ:eq3}
\end{equation}
where $d$ is the offset distance, initialized to zero.According to the standard definition of 2DGS, the $i$-th 2D Gaussian primitive is denoted as:
\begin{equation}
    G(k,u,v,d)=\{\mathbf{p}_i,\mathbf{s}_i,\mathbf{q}_i,\alpha_i,\mathbf{sh}_i\}
\end{equation}
where $\mathbf{s}_i$ denotes the scaling factor, $\mathbf{q}_i$ is the rotation, $\alpha_i$ is the opacity, and $\mathbf{sh}_i \in \mathbb{R}^{B \times 3}$ is the spherical harmonic coefficients used to capture the appearance color. These parameters are initialized as follows: $\mathbf{s}_i$ is set to the distance between the nearest neighboring points, $\mathbf{q}_i = [1, 0, 0, 0]$, $\alpha_i = 0.1$, $\mathbf{sh}_i$ is defined using third-order spherical harmonics with $B=9$, and the barycentric coordinates $(u,v)$ are randomly initialized, and offset $d$ is set to zero. 

We randomly sample a subset of triangles from the SMPL mesh to determine the regions to be embedded with 2D Gaussian primitives. In all experiments, a total of 30,000 triangles are selected. Since joint regions typically undergo significant deformation during motion, we apply a denser sampling strategy in these areas to better preserve their geometric characteristics. Specifically, we leverage the skeletal structure provided by the SMPL model to identify mesh triangles within a predefined distance from the joints, forming a joint-related triangle set denoted as $\mathcal{F}$. Each triangle in $\mathcal{F}$ is individually embedded with a 2D Gaussian primitive, enabling more stable and accurate representation of deformation-sensitive regions.

\subsection{Transformations from Canonical Space to Posed Space}
\label{sec:Transformations}
During training, we apply forward Linear Blend Skinning (LBS) to transform the 2D Gaussian primitives, which are initially embedded in the canonical SMPL mesh, into the posed space corresponding to the $t$-th frame sampled from a monocular video sequence. We first update the center position of each 2D Gaussian primitive by warping the vertices of its associated triangle from the canonical space to the posed space. Next, we compute the rotation of each triangle by analyzing its geometry, including vertex positions and normals in both the canonical and posed spaces~\cite{zielonka2023instant}. To propagate this rotation to mesh vertices, we compute an area-weighted average of the quaternions from all adjacent triangles.

\begin{equation}
    \mathbf{q}_\mathbf{v}(t)=\frac{\sum_{k\in{\mathcal{T}(\mathbf{v})}}A_k^{\text{pose}}(t) \mathbf{q}_k(t)}{\sum_{k\in{\mathcal{T}(\mathbf{v})}}A_k^{\text{pose}}(t)}
    \label{equ:comp_qua}
\end{equation}

where $\mathcal{T}(\mathbf{v})$ denotes the set of triangles adjacent to vertex $\mathbf{v}$, and $A_k^{\text{pose}}(t)$ and $\mathbf{q}_k(t)$ respectively represent the area and the rotation quaternion of the $k$-th triangle in the posed space at frame $t$.

The rigid rotation $\delta \mathbf{q}_{t}$ of each 2D Gaussian primitive is computed by barycentrically interpolating the quaternions of the three vertices of its associated triangle. Meanwhile the scaling factor $\delta \mathbf{s}_{t}$ of each 2D Gaussian primitive is derived as the ratio between the area of the embedded triangle in the posed space and canonical space:
\begin{equation}
    \delta \mathbf{s}_{t}=\frac{A^{pose}}{A^{cano}},\quad\delta \mathbf{q}_{t}=u\mathbf{q}_0+v\mathbf{q}_1+(1-u-v)\mathbf{q}_2
\end{equation}
where $A^{pose}$ and $A^{cano}$ denote respectively the areas of the triangle in the posed space and canonical space at frame $t$. The vertex quaternions ${\mathbf{q}_0, \mathbf{q}_1, \mathbf{q}_2}$ are computed according to Equation~\ref{equ:comp_qua} as area-weighted averages of adjacent triangle rotations. 
\begin{figure}
  \centering
  \includegraphics[width=1\linewidth]{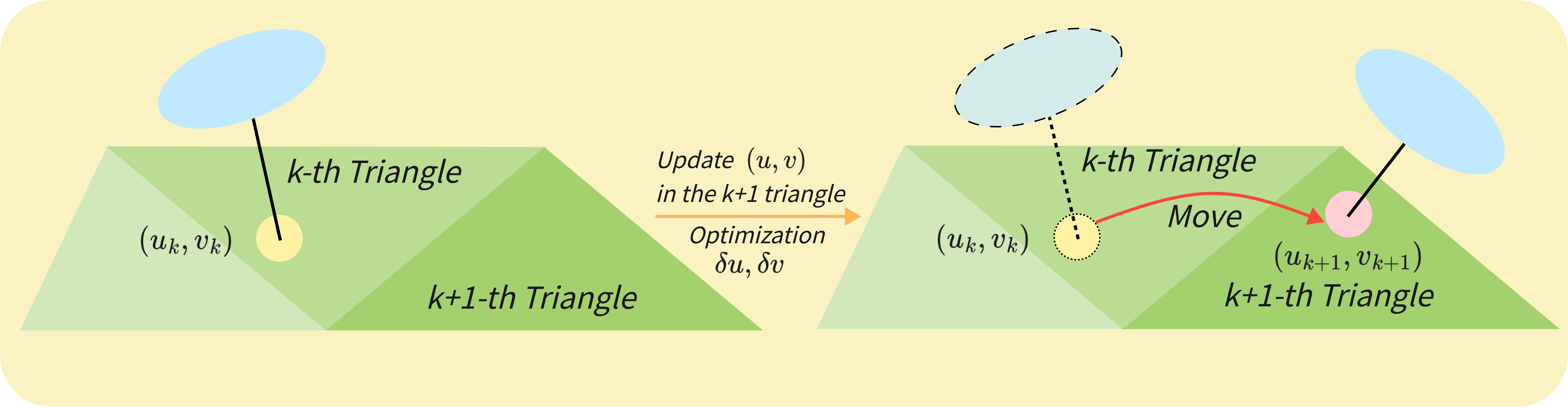}
  \caption{The dynamic update mechanism of the barycentric coordinates for adjacent triangle transfer.} 
  \label{fig:triangle_walk}
\end{figure}
Following Shao \emph{et al.}\cite{shao2024splattingavatar}, during optimization, the barycentric coordinates $(u, v)$ of a 2D Gaussian primitive may shift due to parameter updates. When any of $u$, $v$, or $1{-}u{-}v$ becomes negative, it indicates that the primitive has moved outside the bounds of its current triangle. In such cases, the Gaussian primitive is considered to have crossed into an adjacent triangle, and its barycentric coordinates must be recomputed to reflect the new embedding. As illustrated in Fig.\ref{fig:triangle_walk}, if a 2D Gaussian primitive originally embedded in the $k$-th triangle yields a negative coordinate value after being updated to $(u_k, v_k)$, it indicates the embedding is transferred to the neighboring $(k{+}1)$-th triangle. The local coordinate system is then re-initialized by recalculating valid barycentric coordinates $(u_{k+1}, v_{k+1})$ in the new triangle, ensuring consistent and accurate surface alignment for the human avatar.

\subsection{Rotation Compensation Network}
\label{sec:rotation}
\begin{figure}[t]
  \centering
  \includegraphics[width=1\linewidth]{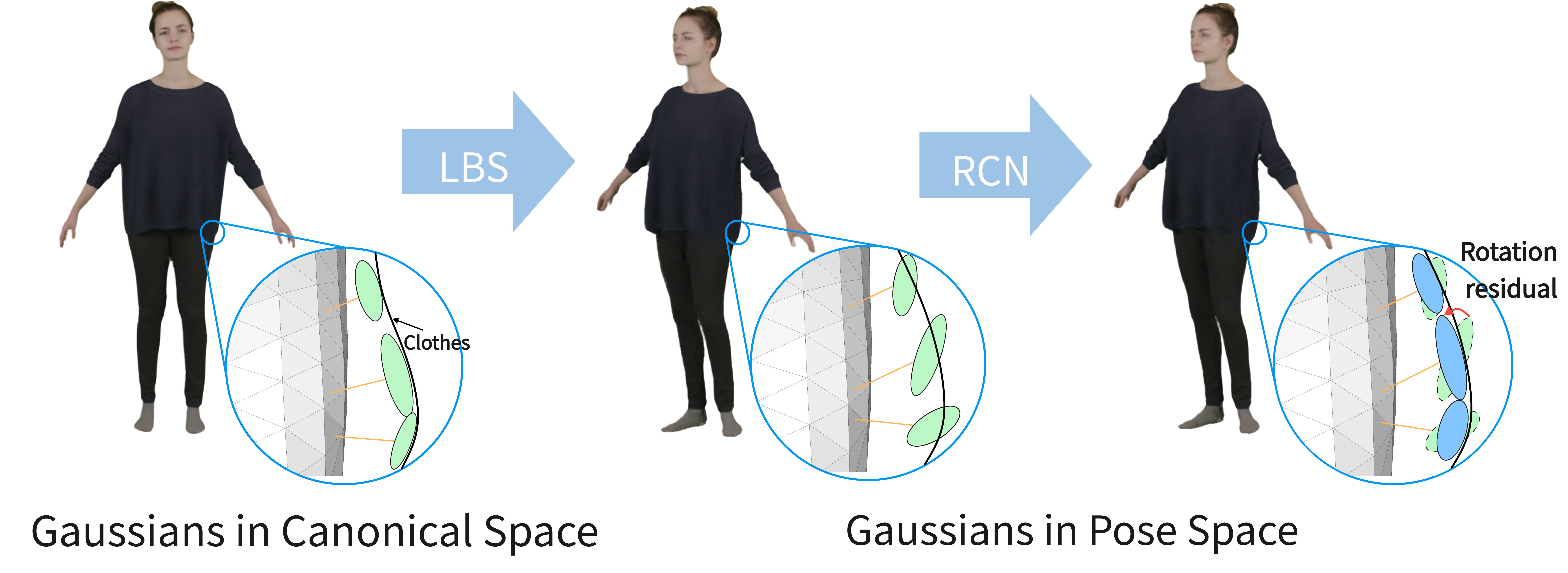}
  \caption{Rigid mesh deformation creates rotation errors (green). RCN learns the rotation residual (red) and corrects it (blue), realigning Gaussians to the true surface.}
  \label{fig:method_rcn}
\end{figure}

In the previous sections, we introduced the use of 2D Gaussian primitives embedded into the SMPL mesh with dense sampling around the joints. These techniques have shown great promise in capturing high-frequency geometric details and improving the overall quality of the reconstructed avatars, particularly in regions with complex surface deformations. However, challenges remain in accurately modeling complex non-rigid deformations and correcting rotation errors, such as the coarse joint rotation interpolation in LBS and the low resolution of the SMPL mesh. These issues lead to inaccuracies in the final avatar, particularly in dynamic regions like clothing and joints as illustrated in the middle of Fig.~\ref{fig:method_rcn}. We introduce the Rotation Compensation Network (RCN), which learns compensation patterns for complex regions through deep learning. By incorporating RCN, the model no longer solely depends on the coarse transformations provided by LBS or the low-resolution localities of the SMPL mesh, thus enabling fine-grained, structurally consistent, and high-fidelity human avatar reconstruction.

To model the rigid rotation residuals of 2D Gaussian primitives in the context of LBS, we introduce a global pose encoder and a local geometry encoder to extract rotation features that depend on both global pose and local surface geometry. These features are then processed by a decoder to compute the rotation compensation for each 2D Gaussian primitive. For rigid rotation compensation of the human avatar, we employ a two-stage training strategy: initially, the avatar is trained using LBS rigid transformations, and once the number of 2D Gaussian primitives stabilizes, the RCN is trained to learn the rotation residuals.

In RCN, each 2D Gaussian primitive is assigned to a specific triangle on the mesh surface, identified by its triangle index $k$ in the SMPL mesh. To incorporate triangle-level semantic information, we embed the triangle index $k$ into a learnable feature vector $\mathbf{f}_k \in \mathbb{R}^{256}$.
To construct the local geometric information for a 2D Gaussian primitive, we concatenate the barycentric coordinates $(u,v)$ (where $u,v \geq 0$ and $u+v \leq 1$), the offset distance $d \in \mathbb{R}$ along the interpolated normal, the triangle embedding $\mathbf{f}_k$, and the 2D Gaussian primitive rotation quaternion $\mathbf{q} \in \mathbb{R}^4$. These components are then fed into a local geometry encoder to extract a high-dimensional local feature vector:
\begin{equation}
    \mathbf{x}_{\text{geo}} = \mathrm{Enc}_{\text{geo}}([u, v, d, \mathbf{f}_k, \mathbf{q}]).
\end{equation}

In parallel, we employ a global pose encoder to capture body-level motion context from the pose parameters $\boldsymbol{\theta} \in \mathbb{R}^{72}$ of the SMPL model. This encoder maps $\boldsymbol{\theta}$ into a compact latent representation:
\begin{equation}
    \mathbf{x}_{\text{pose}} = \mathrm{Enc}_{\text{pose}}(\boldsymbol{\theta}).
\end{equation}

We then concatenate the local and global features and pass them through a residual decoder to predict the rotation residual for each 2D Gaussian primitive:
\begin{equation}
    \mathbf{q}_c = \mathrm{Dec}_{\text{rot}}([\mathbf{x}_{\text{geo}}, \mathbf{x}_{\text{pose}}]),
\end{equation}
Where $\mathbf{q}_c \in \mathbb{R}^4$ is a unit quaternion representing the incremental rotation to be applied to the original orientation. In the posed space, the final rotation $\mathbf{q}_{t}$ and scaling $\mathbf{s}_{t}$ of $i$-th 2D Gaussian primitive are expressed as:
\begin{equation}
    \mathbf{s}_{t}=\delta \mathbf{s}_{t}*{\mathbf{s}_i},\,\mathbf{q}_{t}=\mathbf{q}_{c}\otimes \delta \mathbf{q}_{t}\otimes{\mathbf{q}_i}
    \label{equ:final_rot}
\end{equation}
Where $\otimes$ denotes quaternion multiplication.

This design enables RCN to compensate for the non-rigid deformation effects that cannot be captured by LBS alone, especially in regions with high-frequency surface motion, thereby enhancing the accuracy and realism of the reconstructed human avatar.

\subsection{Loss Functions}
\label{sec:loss}
In this section, we outline the loss functions used to train our model, which are designed to address various aspects of the human avatar reconstruction task. The total loss is composed of several components, each targeting specific challenges, including rotation residuals, scaling regularization, joint region constraints, and surface consistency.

The total loss $L_{all}$ combines multiple terms:$L_1$, $L_{mse}$, $L_{lpips}$, $L_{scaling}$, $L_{joint}$, $L_n$, $L_{rcn}$. The form of the total loss function is as follows: 
\begin{align}
    L_{all} = &\ \lambda_1 L_1 + \lambda_2 L_{mse} + \lambda_3 L_{lpips} \notag \\
             &+ \lambda_4 L_{scaling} + \lambda_5 L_{joint} + \lambda_6 L_n + \lambda_7 L_{rcn}
\end{align}
Each term addresses a specific aspect of reconstruction:

\begin{itemize}
    \item \( L_1 \): Pixel-level accuracy between the reconstructed and ground truth images.
    \item \( L_{\text{mse}} \): Minimizes squared differences between predicted and actual values.
    \item \( L_{\text{lpips}} \): Ensures perceptual similarity between the generated and ground truth images.
    \item \( L_{\text{scaling}} \): Regularizes the scaling of 2D Gaussian primitives to avoid excessive elongation.
    \item \( L_{\text{joint}} \): Applies constraints on scaling near joints to prevent artifacts.
    \item \( L_n \): Enforces normal consistency to ensure smooth surface reconstruction.
    \item \( L_{\text{rcn}} \): Prevents overfitting during novel pose synthesis by controlling rotation compensation.
\end{itemize}
The hyperparameters \( \lambda_1, \lambda_2, \dots, \lambda_7 \) control the relative importance of each loss term, and are empirically set as follows: $\lambda_1=1.0$, $\lambda_2=10.0$, $\lambda_3=0.01$, $\lambda_4=20.0$, $\lambda_5=10.0$, $\lambda_6=0.01$, and $\lambda_7=0.1$. We jointly optimize the parameters of the 2D Gaussian primitives, including scaling \( \mathbf{s} \), rotation \( \mathbf{q} \), opacity $\alpha$, spherical harmonics \( \mathbf{sh} \), barycentric coordinates \( (u, v) \), and offset $d$ along with the weights of the Rotation Compensation Network (RCN). This multi-loss joint optimization framework effectively regularizes the deformation of 2D Gaussian primitives across different regions, ensuring that the reconstructed details remain both faithful to the appearance and geometrically coherent. As a result, it significantly enhances the overall reconstruction quality and robustness of the human avatar.

Due to limitations in monocular video data and the limited pose variations in the training dataset, we follow~\cite{shao2024splattingavatar} and apply scale regularization to the 2D Gaussian primitives to prevent them from becoming excessively elongated during reconstruction. We define $\hat{s}$ and $\check{s}$ as the maximum and minimum principal-axis scaling of the 2D Gaussian primitives, respectively. We define the scale regularization loss as:
\begin{equation}
    L_{\text{scaling}} = \frac{1}{N} \sum_{i\in G} \hat{s}_i
\end{equation}
Where, $G = \{j \mid \hat{s}_i > \epsilon_{s} \quad \text{and} \quad \frac{\hat{s}_i}{\check{s}_i} > \epsilon_{r}\}$,  $\epsilon_{s}$ denotes the predefined maximum scaling threshold, and $\epsilon_{r}$ represents the upper bound of the ratio between the maximum and minimum scaling values. Empirically, we set $\epsilon_s=0.008$ and $\epsilon_r=5.0$. 

To further suppress artifacts in the joint regions, we apply a joint regularization loss that constrains the scaling of the 2D Gaussian primitives near the joints, preventing excessive scaling due to the inherent limitations of monocular data. Specifically, the joint regularization loss is defined as:
\begin{equation}
    \label{equ:joint_limi}
    L_{joint}=\frac{1}{N}\sum_{i\in\{j|max(s_{j,u},s_{j,v})>\tau\}}\max(s_{i,u},s_{i,v})
\end{equation}
Where $s_{i,u}$ and $s_{i,v}$ denote the scaling factors of the $i$-th 2D Gaussian primitive along the two principal axes within the joint region set $\mathcal{F}$, and $\tau$ denotes the predefined scaling threshold. The joint area is determined using the skeletal priors of the SMPL model.

To enhance the surface details of human avatar reconstruction, we employ a normal consistency loss~\cite{huang20242d} that encourages the 2D Gaussian primitives to align with the surface normals. This loss is defined as:
\begin{equation}
    \label{equ:normal_depth}
    L_\mathrm{n}=\sum_\mathrm{i}\mathrm{~w_i(1-\mathbf{n}_i^T\mathbf{N})}
\end{equation}
Where $\mathrm{w}_i$ is the blending weight for the $i$-th intersection, $\mathbf{n}_i$ is the normal at the $i$-th intersection, and $\mathbf{N}$ is the normal derived from the depth map.

To address the rigid rotation residuals introduced by LBS, we introduce a rotation compensation loss $L_{rcn}$. While LBS effectively models rigid transformations, it fails to capture non-rigid deformations, particularly in areas such as joints and clothing. The rotation compensation loss refines the rigid transformations by compensating for inaccuracies introduced by LBS, ensuring that the predicted rotations align more closely with the real surface of the human avatar.The loss is computed as:
\begin{equation} L_{rcn}=\left|1-\frac{1}{N}\sum_{i=1}^N\langle\mathbf{q}_{lbs}^{(i)}, \mathbf{q}_{rcn}^{(i)}\rangle\right| 
\end{equation} 
Where $\mathbf{q}_{lbs}^{(i)}$ is the rotation of the $i$-th 2D Gaussian primitive after rigid deformation, $\mathbf{q}_{rcn}^{(i)}$ is the rotation after compensation, $\langle\cdot\rangle$ denotes the quaternion dot product. This loss improves the accuracy and realism of the human avatar, particularly in handling complex pose variations, by compensating for the non-rigid transformations that LBS cannot model.

\section{Experiments}
\label{sec:exp}
\subsection{Datasets and Metrics}
\noindent\textbf{Datasets.}
The test data is selected from the publicly available PeopleSnapshot\cite{alldieck2018detailed} and Synthetic\cite{jiang2022selfrecon} datasets, with four human video sequences chosen from each dataset. Videos in both datasets were captured using a static monocular camera, recording full 360° rotations of subjects in the standard A-pose. For the PeopleSnapshot dataset, we strictly follow the experimental protocol of InstantAvatar~\cite{jiang2023instantavatar} for data partitioning. For the Synthetic dataset, we apply the same training strategy and testing split criteria to ensure consistent and fair comparisons across methods.

\noindent\textbf{Evaluation metrics.}
To comprehensively evaluate the proposed method on novel view synthesis, we conduct quantitative analysis using a set of three complementary metrics: Peak Signal-to-Noise Ratio (PSNR), Structural Similarity Index (SSIM), and Learned Perceptual Image Patch Similarity (LPIPS)~\cite{Zhang_Isola_Efros_Shechtman_Wang_2018}. PSNR quantifies pixel-level discrepancies between synthesized avatars and ground-truth images, SSIM evaluates structural similarity, and LPIPS focuses on perceptual-level discrepancies, providing a comprehensive evaluation of low-level accuracy, structural fidelity, and perceptual quality.

\subsection{Experimental Settings}
\noindent\textbf{Implementation details.} We employ a two-phase training strategy. Initially, the model is trained with LBS transformations to stabilize the configuration of the 2D Gaussian primitives. Once stabilized, we train the RCN to learn and compensate for the rotation residuals introduced by LBS. This staged approach ensures that the model gradually refines rotation compensation, leading to more accurate and realistic human avatar reconstructions. At each iteration, we randomly select a single training frame as input. The model is optimized using the Adam optimizer~\cite{kingma2014adam} with an initial learning rate of $5\times 10^{-4}$. In the first stage, we perform 30,000 iterations to optimize the 2D Gaussian avatar, followed by an additional 10,000 iterations in the second stage to fine-tune the RCN. All experiments are conducted on a single NVIDIA GeForce RTX 3090 GPU, with the entire training process taking approximately 25 minutes to complete.

 \noindent\textbf{Methods for comparision.} We compare our method against three state-of-the-art human avatar reconstruction approaches: GaussianAvatar\cite{hu2024gaussianavatar}, Gart\cite{lei2024gart}, and SplattingAvatar~\cite{shao2024splattingavatar}. All comparison methods are implemented using their most recent publicly available code and are trained on the same test dataset to ensure fairness in evaluation.

\label{sec:novel_views}
\begin{figure*}
  \centering
  \includegraphics[width=1\linewidth]{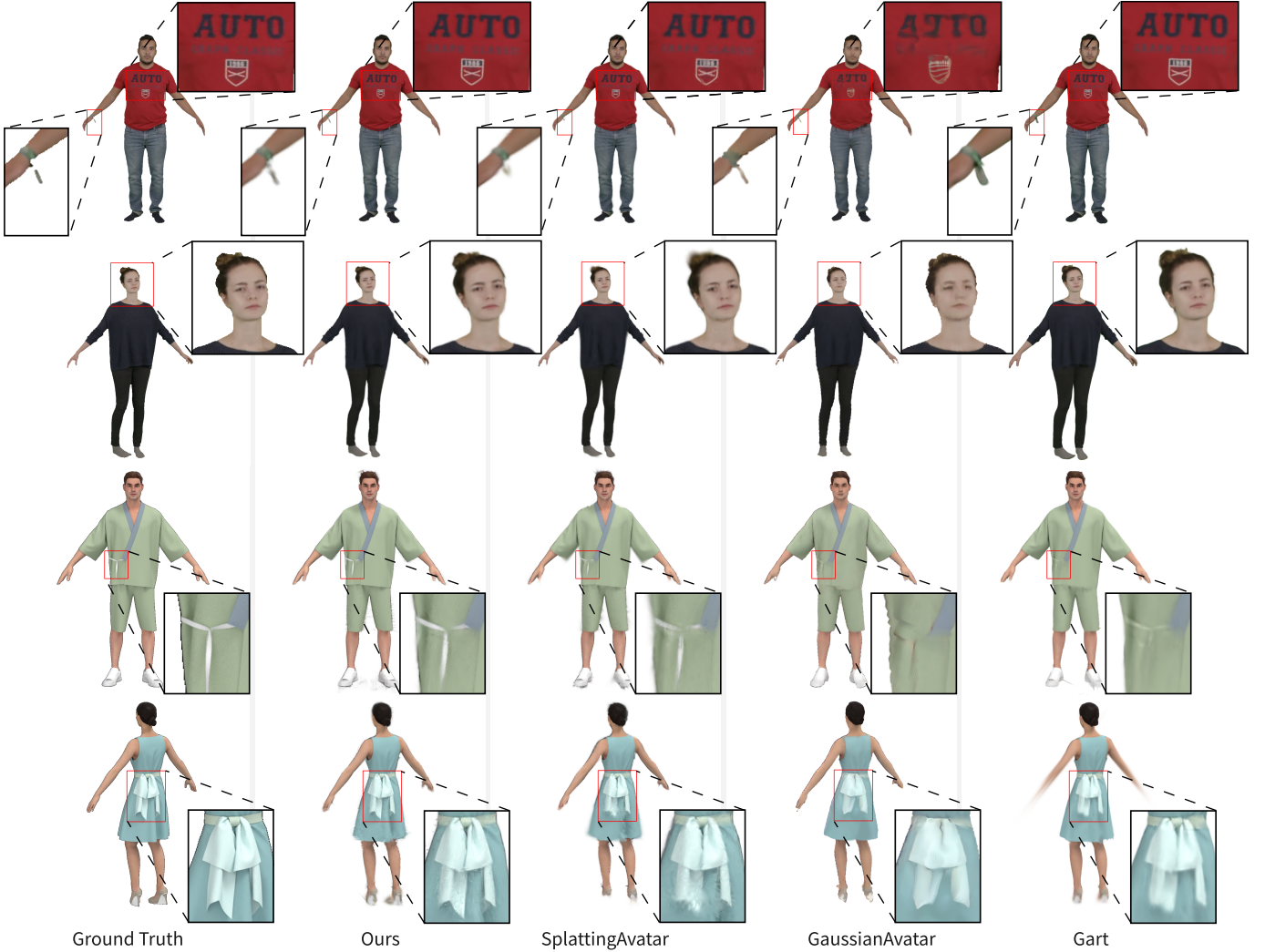}
  \caption{Comparisons on the PeopleSnapshot~\cite{alldieck2018detailed} and Synthetic~\cite{jiang2022selfrecon} datasets. Rows 1–2 show reconstruction results on the PeopleSnapshot dataset, and Rows 3–4 present comparisons on the Synthetic dataset. From left to right, the columns correspond to Ground Truth, our method, SplattingAvatar~\cite{shao2024splattingavatar}, GaussianAvatar~\cite{hu2024gaussianavatar}, and Gart~\cite{lei2024gart}.}
  \label{fig:novel_views}
\end{figure*}

\subsection{Comparisons on Novel Views Synthesis}

\begin{table*}[!t]
    \caption{Quantitative comparisons on PeopleSnapshot~\cite{alldieck2018detailed} and Synthetic Dataset~\cite{jiang2022selfrecon}.}
    \centering
    \begin{tabular}{lcccccccccccc}
    \hline
    \multicolumn{13}{c}{PeopleSnapshot} \\
                    & \multicolumn{3}{c}{male-3-casual} & \multicolumn{3}{c}{male-4-casual} & \multicolumn{3}{c}{female-3-casual} & \multicolumn{3}{c}{female-4-casual} \\
                    & PSNR$\uparrow$ & SSIM$\uparrow$ & LPIPS$\downarrow$ & PSNR$\uparrow$ & SSIM$\uparrow$ & LPIPS$\downarrow$ & PSNR$\uparrow$ & SSIM$\uparrow$ & LPIPS$\downarrow$ & PSNR$\uparrow$ & SSIM$\uparrow$ & LPIPS$\downarrow$ \\ \hline
    GaussianAvatar  & 30.98 & 0.979 & \textbf{0.015} & 28.78 & 0.976 & \textbf{0.023} & 29.55 & 0.976 & \textbf{0.023} & 30.84 & 0.977 & \textbf{0.014} \\
    GART            & 29.95 & 0.978 & 0.038        & 27.20 & 0.968 & 0.061        & 25.29 & 0.963 & 0.049        & 28.34 & 0.971 & 0.037        \\
    SplattingAvatar & 32.30 & 0.978 & 0.031        & 30.51 & 0.979 & 0.041        & 30.45 & 0.977 & 0.043        & 32.21 & 0.976 & 0.032        \\
    Ours             & \textbf{33.92} & \textbf{0.984} & 0.020 & \textbf{31.25} & \textbf{0.982} & 0.030 & \textbf{31.37} & \textbf{0.980} & 0.031 & \textbf{33.22} & \textbf{0.982} & 0.020 \\ \hline
    \multicolumn{13}{c}{Synthetic Data} \\
                    & \multicolumn{3}{c}{female1} & \multicolumn{3}{c}{female2} & \multicolumn{3}{c}{male1} & \multicolumn{3}{c}{male2} \\
                    & PSNR$\uparrow$ & SSIM$\uparrow$ & LPIPS$\downarrow$ & PSNR$\uparrow$ & SSIM$\uparrow$ & LPIPS$\downarrow$ & PSNR$\uparrow$ & SSIM$\uparrow$ & LPIPS$\downarrow$ & PSNR$\uparrow$ & SSIM$\uparrow$ & LPIPS$\downarrow$ \\ \hline
    GaussianAvatar  & 24.04 & 0.943 & \textbf{0.062} & 25.04 & 0.949 & \textbf{0.059} & 24.57 & 0.937 & \textbf{0.067} & 25.24 & 0.946 & \textbf{0.057} \\
    GART            & 23.04 & 0.938 & 0.140        & 25.56 & 0.953 & 0.127        & 25.22 & 0.944 & 0.128        & 25.65 & 0.951 & 0.113        \\
    SplattingAvatar & 25.31 & 0.949 & 0.109 & 26.21 & \textbf{0.957} & 0.109 & 25.77 & 0.945 & 0.118 & 26.25 & \textbf{0.954} & 0.107 \\
    Ours             & \textbf{26.53} & \textbf{0.952} & 0.097 & \textbf{26.55} & 0.954 & 0.103 & \textbf{26.81} & \textbf{0.946} & 0.103 & \textbf{27.16} & \textbf{0.954} & 0.095 \\ \hline
    \end{tabular}
    \label{tab:exp_all}
\end{table*}

The novel view synthesis task primarily evaluates the quality of reconstructed human avatars by generating images from previously unseen viewpoints in the canonical space. As shown in Fig.~\ref{fig:novel_views}, we compare our method with three state-of-the-art methods. SplattingAvatar fits clothing by adding offsets to 3D Gaussian primitives. However, these offsets are optimized solely based on photometric loss without geometric constraints, causing the primitives to misalign with the true body surface. As a result, fine details such as the hair accessory in the second row and the waist belts in the third and fourth rows appear blurred or missing. GaussianAvatar adopts isotropic Gaussian primitives to maintain view-consistent appearance, but their nearly spherical shape limits their ability to capture surface details. Consequently, artifacts such as the distorted logo in the first row and the missing belt in the third row are observed. Gart models dynamic body deformation by binding Gaussian primitives to skeletal joints. Yet it occasionally misassigns joints, leading to missing body parts. For example, the female character’s hand in the fourth row. Moreover, its volumetric Gaussian representation, which models full angular radiance, is incompatible with the thin structure of human surfaces, making it difficult to reconstruct fine geometry like the belt and bow. In contrast, our method achieves more accurate reconstruction of surface details. This is due in part to the use of surface normal and depth consistency constraints, which guide Gaussian primitives to better adhere to the true body surface. In addition, the proposed RCN further refines their orientation, enabling precise reconstruction of fine structures such as the belt and the bow, as demonstrated in the third and fourth rows.

We further perform a quantitative evaluation to compare our method with the three comparison approaches in novel view synthesis. As demonstrated in Tab.~\ref{tab:exp_all}, our method achieves the highest PSNR scores across all test cases, outperforming the second-best method, SplattingAvatar, by an average margin of 0.975. This is mainly attributed to the more accurate reconstruction of human body details by our method, which leads to higher pixel-level reconstruction quality in the synthesized novel-view images. In terms of SSIM, all methods yield very similar scores, with our approach achieving slightly higher values than the other three methods, which indicates that our method performs better in preserving human structural integrity during reconstruction.For the LPIPS metric, our method achieves a reduction of $28\%$ and $15\%$ compared to Gart and SplattingAvatar, respectively, demonstrating its superiority in generating perceptually more realistic human images. While GaussianAvatar attains a relatively low LPIPS score by adopting isotropic Gaussian distributions and a simplified optimization strategy, this design choice comes at the cost of reduced expressiveness in modeling fine-grained local details. Consequently, the surface textures of the reconstructed avatars exhibit an 'averaging' effect, which adversely impacts its performance on pixel-level metrics such as PSNR.

\subsection{Comparisons on Novel Pose Synthesis}
\label{sec:novel_poses}
\begin{figure}[t]
  \centering
  \includegraphics[width=1\linewidth]{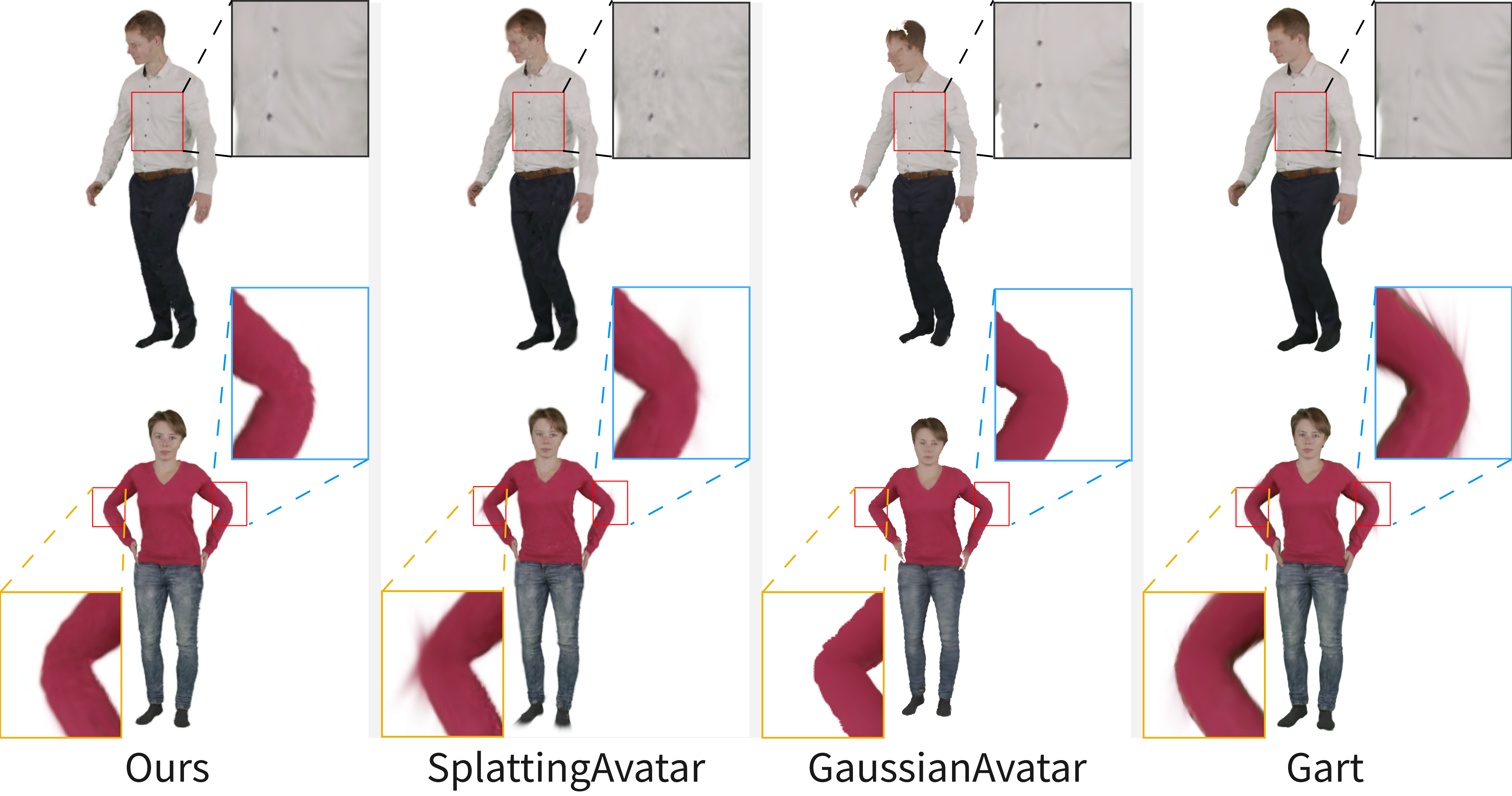}
  \caption{Qualitative comparison of our method with three state-of-the-art methods under novel pose conditions. From left to right: our method, SplattingAvatar~\cite{shao2024splattingavatar}, GaussianAvatar~\cite{hu2024gaussianavatar}, and Gart~\cite{lei2024gart}.}
  \label{fig:novel_poses}
\end{figure}

The novel pose synthesis task primarily involves modifying the SMPL pose parameters to animate the human avatar into new poses, while simultaneously enabling novel-view rendering under these poses. For evaluation, we compare our method with three comparison methods by synthesizing images corresponding to the same set of target poses. 

In SplattingAvatar, during human motion animation, the transformation of 3D Gaussian primitives is solely driven by the rigid motions, namely rotation and translation, of the associated SMPL mesh triangles. This rigid coupling fails to account for complex non-rigid deformations in articulated body regions, resulting in suboptimal surface alignment and noticeable artifacts. As illustrated in the second row of Fig.~\ref{fig:novel_poses}, the Gaussian primitives in joint regions appear detached from the body surface, resulting in visible artifacts. GaussianAvatar~\cite{hu2024gaussianavatar} effectively eliminates artifacts on the body surface, it predicts the position, rotation, and scale of each primitive using a multi-layer perceptron (MLP). However, due to the absence of explicit physical constraints, the MLP-predicted parameters may exhibit errors, resulting in the omission of certain body regions. As shown in the first row of Fig.~\ref{fig:novel_poses}, the head of the male avatar contains a large missing area. Gart~\cite{lei2024gart} introduces additional latent joints to better handle non-rigid deformations and improves synthesis quality across diverse poses, it remains constrained by the fundamental limitations of 3D Gaussian Splatting (3DGS) in modeling fine surface details. Therefore, certain structures, such as the shirt buttons on the male avatar in the first row of Fig.~\ref{fig:novel_poses}, are entirely missing. In contrast, our method leverages geometric constraints to ensure that 2D Gaussian primitives are smoothly distributed along the avatar surface. Furthermore, the proposed RCN module learns to compensate for rotational deviations, effectively mitigating distortions of the 2D Gaussian primitives. Consequently, our method achieves superior rendering quality in both clothing details and articulated joint regions compared to the comparison methods.

\begin{figure}[t]
  \centering
  \includegraphics[width=1\linewidth]{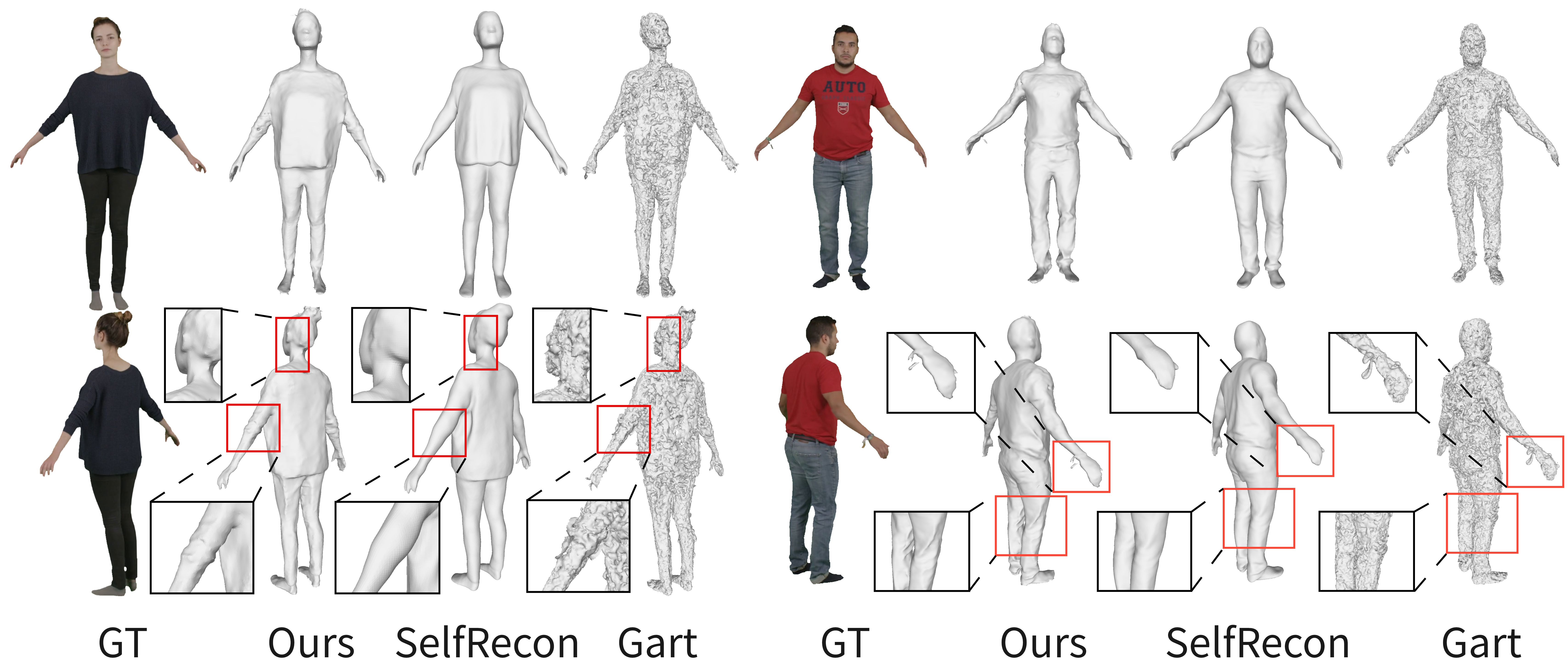}
  \caption{Comparisons of mesh reconstruction results among our method, SelfRecon~\cite{jiang2022selfrecon}, and Gart~\cite{lei2024gart}.}
  \label{fig:mesh}
\end{figure}

\subsection{Results on mesh reconstruction}
\label{sec:mesh}

To further verify the accuracy of the reconstructed surface geometry, We employ a signed distance function (SDF) to convert the human avatars generated by our method into mesh representations on the PeopleSnapshot~\cite{alldieck2018detailed} dataset. We then compare the mesh reconstruction quality with Gart~\cite{lei2024gart}, a previously introduced 3DGS-based human avatar reconstruction method, as well as SelfRecon~\cite{jiang2022selfrecon}, a representative method specifically designed for detailed human mesh recovery. The comparison results are shown in Fig.~\ref{fig:mesh}. To enhance the quality of the mesh converted from Gart's reconstruction results, we first apply RepKPU super-resolution algorithm~\cite{rong2024repkpu} to refine the geometry of the original point cloud. Subsequently, Poisson surface reconstruction~\cite{kazhdan2006poisson} is employed to recover a high-fidelity surface, followed by mesh extraction using the marching cubes algorithm~\cite{lorensen1998marching}. Despite these post-processing steps, as illustrated in the last column of Fig.~\ref{fig:mesh}, the reconstructed mesh still exhibits significant surface noise due to the geometric approximation errors inherent in the 3DGS representation. SelfRecon~\cite{jiang2022selfrecon} demonstrates strong performance in modeling non-rigid deformations, such as clothing wrinkles. However, as a method specifically tailored for mesh reconstruction, it requires up to 24 hours of training. In contrast, our approach achieves comparable overall reconstruction quality with only 1/48 of the training time (i.e., 30 minutes). Furthermore, our method exhibits superior fidelity and completeness in reconstructing fine-grained details, such as the female avatar’s ears and the male avatar’s wrist accessories.

\subsection{Ablation Studies}
\label{sec:abla}
To validate the influence of the key components and loss function of the proposed model on the quality of reconstruction, we test $3$ model variants on the test set.
\begin{itemize}
    \item Model A~(w/o RCN) is constructed by removing RCN from the full model to evaluate the effectiveness of the proposed primitive rotation compensation strategy.
    \item Model B~(w/o Joint) is constructed by removing the joint constraints from the full model, including the dense sampling strategy around joint regions and the joint regularization loss function.
    \item Model C~(w/o Scale) is trained without the joint regularization loss and the scale regularization loss to evaluate the effectiveness of the scale constraints applied to the 2D Gaussian primitives.
\end{itemize}

\begin{table}[t]
  \caption{Quantitative results of the ablation studies.}
  \centering
  \begin{tabular}{cccc}
    \hline
                         & PSNR↑ & SSIM↑ & LPIPS↓ \\ 
    \hline
    Full Model           & \textbf{32.44} & \textbf{0.982} & \textbf{0.0251}  \\
    w/o RCN             & 32.02 & 0.981 & 0.0265 \\
    w/o Joint            & 32.39 & 0.982 & 0.0252  \\
    w/o Scale            & 32.31 & 0.982 & 0.0256 \\
    \hline
  \end{tabular}
\label{tab:absolute}
\end{table}

Table~\ref{tab:absolute} presents the quantitative evaluation results of the ablation studies.It can be observed that incorporating RCN improves the PSNR and LPIPS metrics of the synthesized human images by $1.3\%$ and $5.6\%$, respectively, indicating that RCN helps enhance the quality of the generated human avatars at both the pixel and perceptual levels. The relatively small improvement is mainly due to that RCN focusing on refining fine details, such as clothing wrinkles and joint deformations, which cover only a small part of the human body and therefore have limited impact on the overall evaluation metrics.Similarly, although the joint constraints and scale constraints contribute to improvements in PSNR and LPIPS, the extent of enhancement is relatively limited. This is mainly because these components and constraints are designed to handle highly deformable regions like joints, which occupy a relatively small portion of the body and thus have a limited effect on the final scores. For the SSIM metric, all three model variants show little to no effect. This suggests that under the guidance of the SMPL model, the reconstructed human avatars can still maintain correct structural information, even without additional refinements or constraints.

\begin{figure}[t]
\centering
\includegraphics[width=1\linewidth]{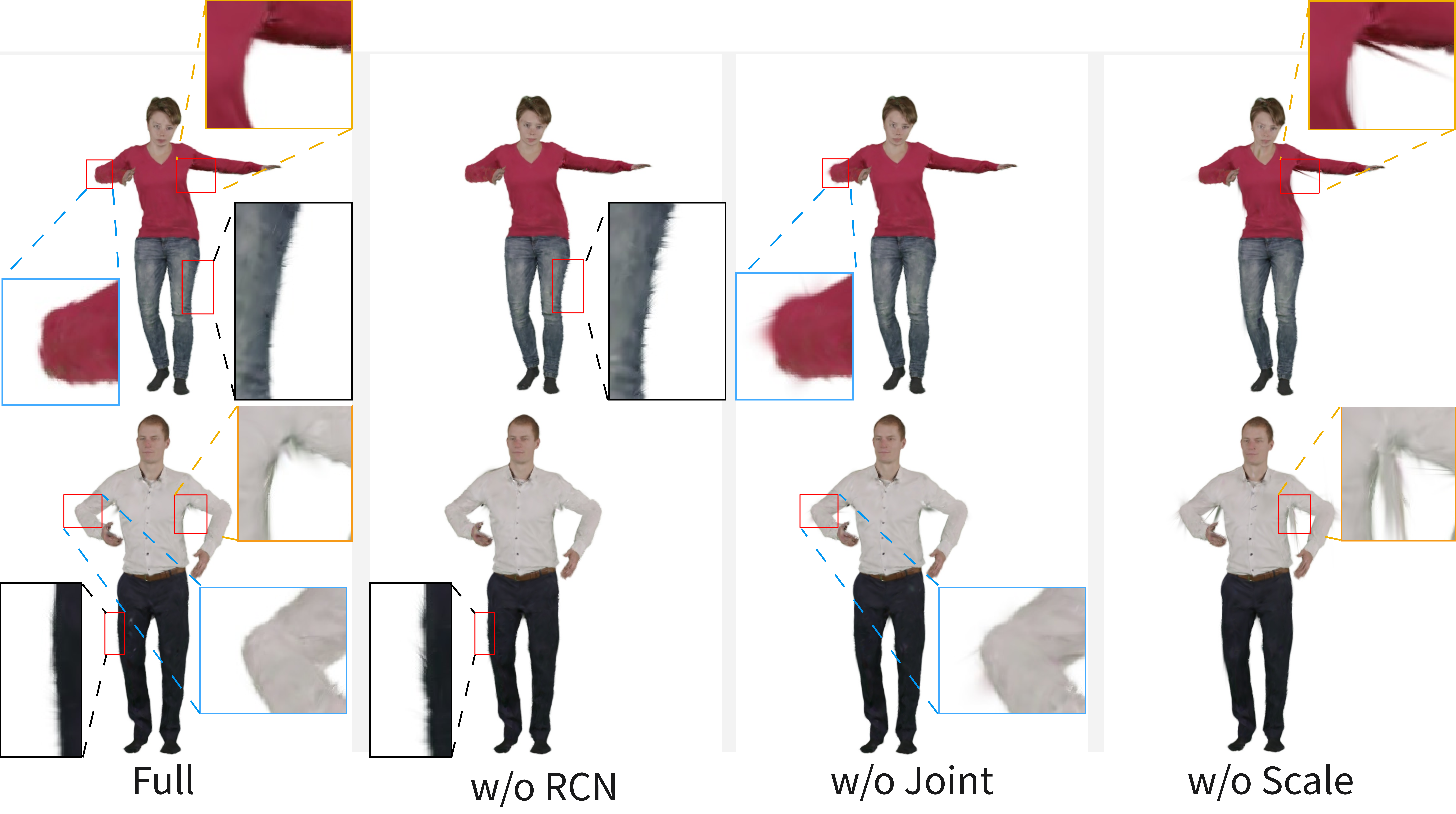}
\caption{Visual comparison of novel pose synthesis results for human avatars reconstructed using the full model and the three model variants.}
\label{fig:absolute}
\end{figure}

\begin{figure}[t]
\centering
\includegraphics[width=1\linewidth]{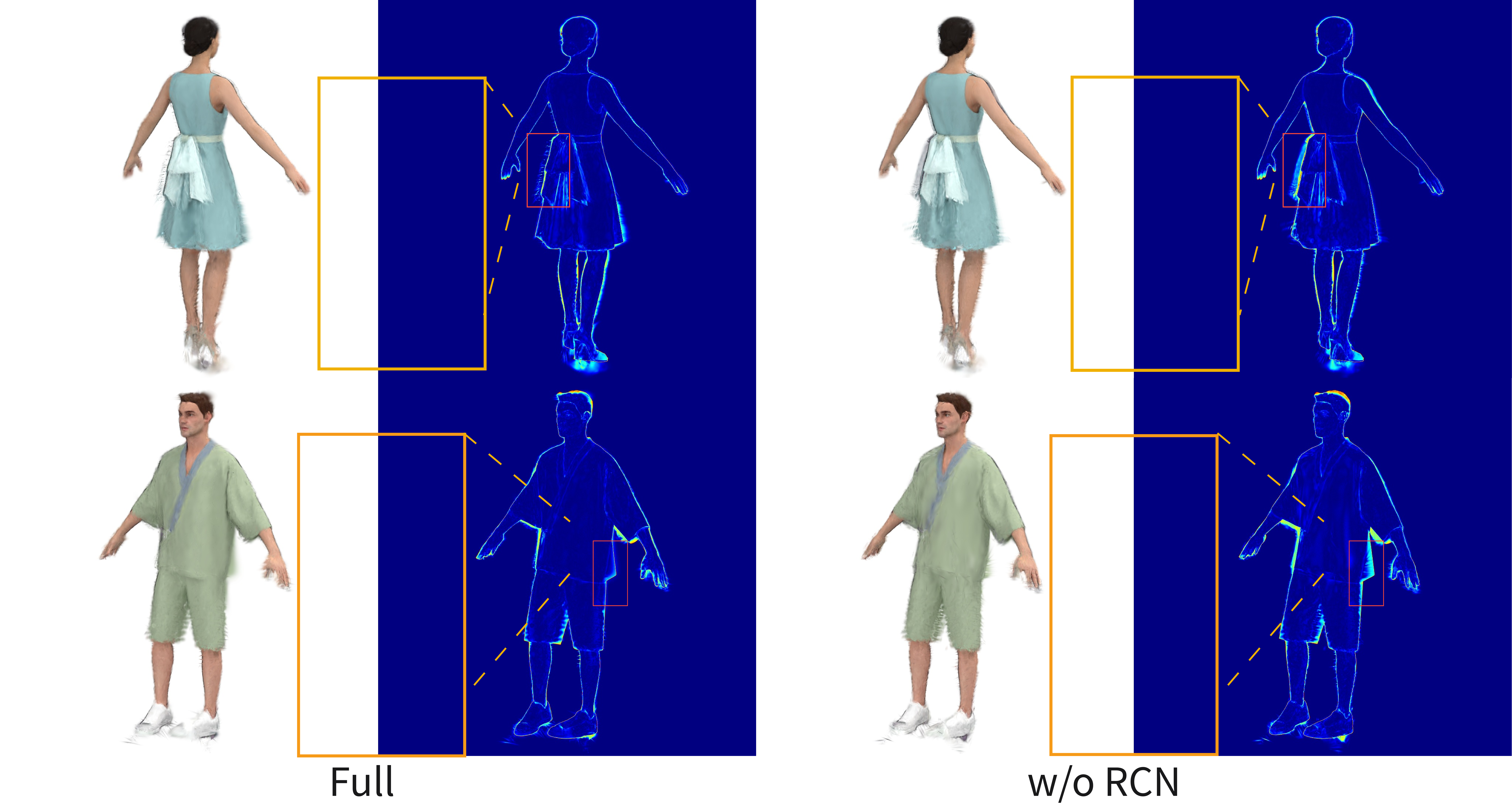}
\caption{Visual comparison of reconstruction errors in human images generated with and without using RCN.} 
\label{fig:absolute_RCN}
\end{figure}

\begin{figure}[t]
\centering
\includegraphics[width=1\linewidth]{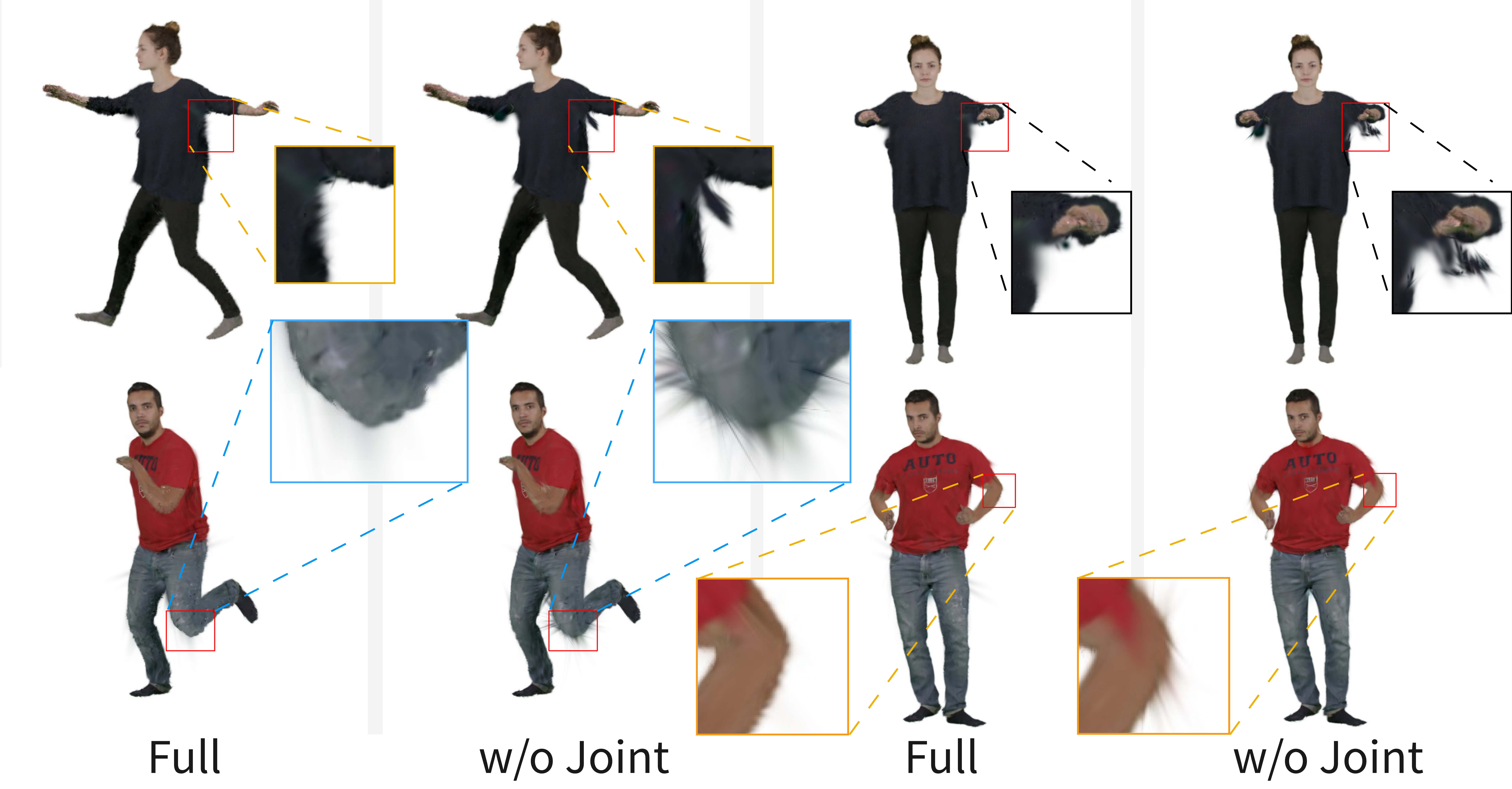}
\caption{Visual comparison with and without using joint constraints. }
\label{fig:absolute_joint}
\end{figure}

\begin{figure}[t]
\centering
\includegraphics[width=1\linewidth]{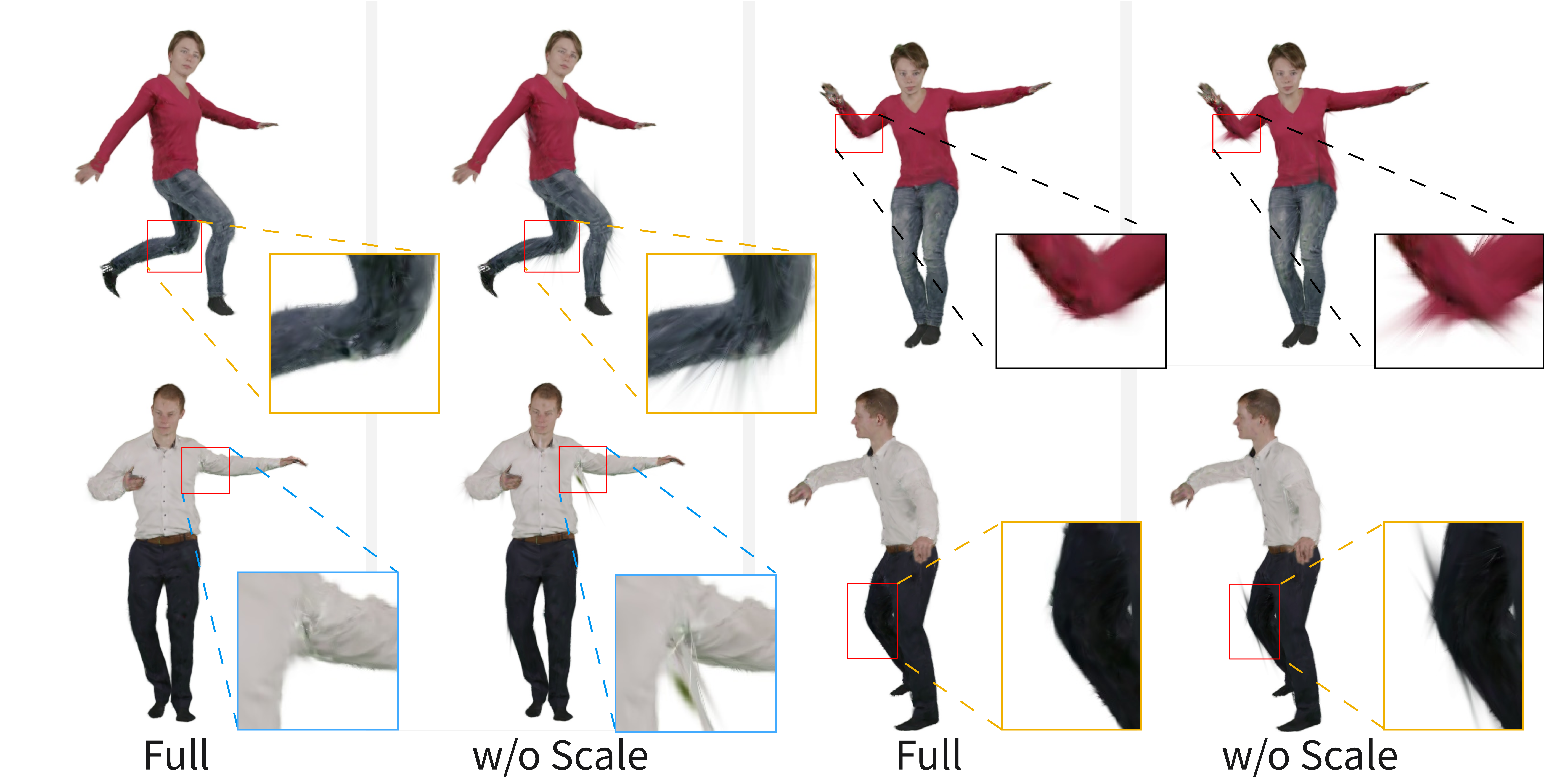}
\caption{Visual comparison with and without using scale constraints. }
\label{fig:absolute_scale}
\end{figure}

Figure~\ref{fig:absolute} presents a visual comparison of novel-pose human images generated by the complete model and its three simplified variants. It is clear that all three variants result in a noticeable decline in the quality of the synthesized images. Specifically, removing the RCN leads to more prominent artifacts around the body boundaries and joints. Figure~\ref{fig:absolute_RCN}, which shows the $l_1$ error maps of the generated results, further confirms this observation by illustrating significantly larger errors along the body boundaries when the RCN is removed. This indicates that the RCN effectively corrects rotation errors of the 2D Gaussian primitives, helping them better align with the human surface and thereby reducing errors in detail-sensitive areas such as boundaries and joints. The third column in Figure~\ref{fig:absolute} and Figure~\ref{fig:absolute_joint} illustrates the results without using joint constraints. We can find that the synthesized images contain numerous artifacts around the joints, with many parts visibly protruding from the body surface. This suggests that the proposed joint-handling strategy enhances the ability of the 2D Gaussian primitives to adapt to regions undergoing large non-rigid deformations during motion. Finally, when the scale of the 2D Gaussian primitives is left unconstrained, as shown in the last column of Figure~\ref{fig:absolute} and in Figure~\ref{fig:absolute_scale}, visible bulges appear in areas such as the armpits, knees, and elbows. This is mainly due to the fact that, without scale constraints, certain Gaussian primitives become excessively elongated. When these elongated primitives lie in regions with large deformations during motion, they tend to extend beyond the human surface, resulting in abnormal bulging artifacts.

\begin{figure}[t]
\centering
\includegraphics[width=1\linewidth]{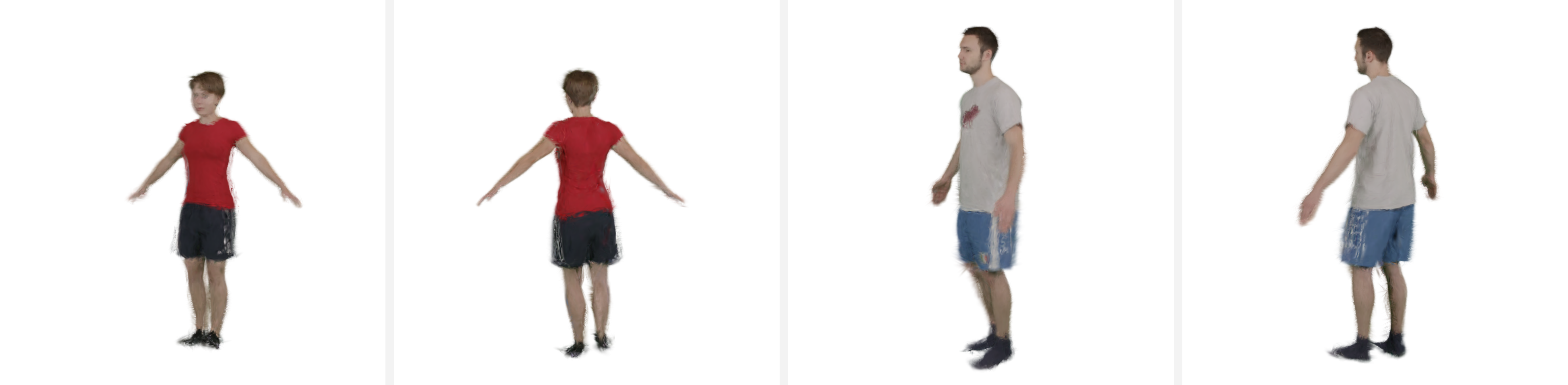}
\caption{Failure cases of our method.}
\label{fig:limitation}
\end{figure}

\subsection{Limitations}
\label{sec:limitation}
Since the SMPL mesh of the human body serves as the foundation for driving the position and shape deformation of 2D Gaussian primitives, the quality of the reconstructed human avatar in this work heavily depends on the accuracy of human pose estimation of the training data. Significant errors in pose estimation within the training data lead to noticeable defects on the surface and along the boundaries of the reconstructed human avatars as illustrated in Figure~\ref{fig:limitation}. Moreover, the overall appearance of the reconstructed human body becomes considerably blurred. This strong dependency on accurate human pose estimation remains a fundamental challenge for current research on monocular video-based human avatar reconstruction.

\section{Conclusion}
\label{sec:conclusion}
In this paper, we propose a new real-time method for building animatable human avatars from monocular video, using 2D Gaussian Splatting (2DGS). This approach supports high-quality image generation of human bodies under novel views and poses. We achieve motion-driven deformation by mapping 2D Gaussian primitives to the SMPL model using Linear Blend Skinning (LBS), and further improve the alignment by introducing a Rotation Compensation Network (RCN) to handle non-rigid body movements. To reduce visual artifacts around joints in new poses, we also design a joint regularization mechanism, which improves animation stability. Extensive experiments demonstrate that our method achieves state-of-the-art performance on both novel view and novel pose synthesis tasks, outperforming existing approaches.

\bibliographystyle{IEEEtran}
\bibliography{ref_ieee_named_abbr}

\end{document}